\documentclass[
]{ceurart}

\usepackage{listings}

\usepackage{makecell}
\usepackage{subcaption}

\definecolor{dkgreen}{rgb}{0,0.6,0}
\definecolor{gray}{rgb}{0.5,0.5,0.5}
\definecolor{mauve}{rgb}{0.28,0,0.42}
 
\begin{document}

\copyrightyear{2023}
\copyrightclause{Copyright for this paper by its authors.
  Use permitted under Creative Commons License Attribution 4.0
  International (CC BY 4.0).}

\conference{AIC 2023: 9\textsuperscript{th} International Workshop on Artificial Intelligence and Cognition,
  September 14--15, 2023, Bremen, Germany}

\title{Formalising Natural Language Quantifiers for Human-Robot Interactions}

\author[1]{Stefan Morar}[%
orcid=0009-0006-5818-1221,
email=stefan.morar30@gmail.com,
]
\address[1]{Department of Computer Science, Technical University of Cluj-Napoca, Romania}

\author[1]{Adrian Groza}[%
orcid=0000-0003-0143-5631,
email=adrian.groza@cs.utcluj.ro,
]

\author[2]{Mihai Pomarlan}[%
orcid=0000-0002-1304-581X,
email=pomarlan@uni-bremen.de,
]
\address[2]{Department of Linguistics, University of Bremen, Germany}


\begin{abstract}
  We present a method for formalising quantifiers in natural language in the context of human-robot interactions. The solution is based on first-order logic extended with capabilities to represent the cardinality of variables, operating similarly to generalised  quantifiers. To demonstrate the method, we designed an end-to-end system able to receive input as natural language, convert it into a formal logical representation, evaluate it, and return a result or send a command to a simulated robot.
\end{abstract}


\maketitle

\section{Introduction}
The translatation from natural language (NL) into logical form and the following transformation into robotic actions is gaining more and more attention~\cite{LONGO202364}.
While human-robot collaboration is a domain continuously evolving
\cite{LIU2023104294}, the NLU models continue to face challenges in accurately capturing the semantics of quantifiers~\cite{cui-etal-2022-generalized-quantifiers}. The efficient representation of different types of quantifiers (e.g. \textit{all}, \textit{most}, \textit{many}, \textit{several}) remains a topic which offers ample opportunity for further enhancement. Several popular logical representations of meaning in natural language processing, such as~\cite{banarescu-etal-2013-abstract}, ignore quantifiers.

This paper focuses on the challenge of effectively managing quantifiers within human-robot interactions and converting natural language instructions into robot  behaviour. We have developed an NLU module which is able to receive 2 types of sentences involving quantifiers: (i) queries, used to gather information regarding the current state of the environment, and (ii) commands, designed to direct the behaviour of a robot. All sentences are converted into logic using an OpenAI GPT-3 model~\cite{finetuning}, fine-tuned with a language dataset created for mapping between English and logic. Using Mace4~\cite{mace4}, a program specialised in evaluating FOL and finding models (i.e. interpretations) that satisfy the given expressions, the FOL expressions generated by the fine-tuned GPT-3 model are evaluated against a knowledge base. Following this evaluation, the system either provides informative responses or transmits appropriate commands to the robot.

The NLU module is integrated with a simulator (AbeSim), specifically designed for a kitchen environment. While the current NLU solution accepts queries and commands related to the cooking domain, the presented approach can be extended to other topics too.

This article is organised as follows: Section~\ref{section:architecture} presents the system architecture, Section~\ref{section:logical-representation} describes the logical representation of natural language (NL) sentences, and Section~\ref{section:qunatifiers} introduces the quantifiers supported by the NLU module. Section~\ref{section:reasoning} describes the FOL evaluator, and Section~\ref{section:abe} the AbeSim simulator. In Sections~\ref{section:fine-tuning} and~\ref{section:results}, we present the fine-tuning methodology and results. Finally, in Section~\ref{section:conclusion}, we draw conclusions and discuss potential areas for improvement.

\section{System Architecture}
\label{section:architecture}

\begin{figure}
\centering
\includegraphics[width=0.65\linewidth]{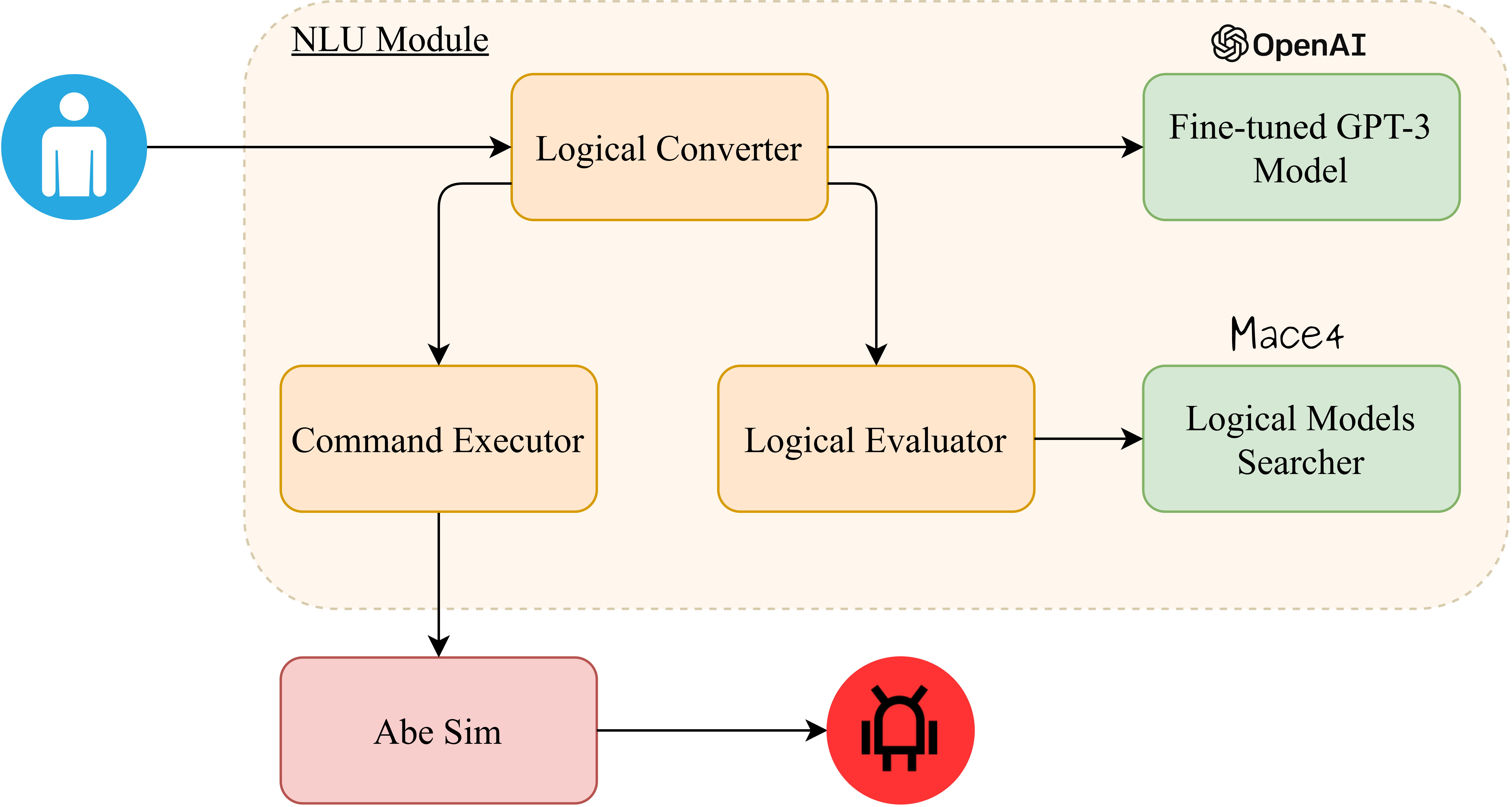}
\caption{System architecture: input in NL is formalised in FOL, interpreted, and given to the AbeSim simulator}
\label{fig:architecture}
\end{figure}

At the top level, there are two modules (Figure~\ref{fig:architecture}): 
(i) the NLU module, which transforms and interprets sentences, and 
(ii) the AbeSim simulator, which executes commands and interacts with the robot. 
The NLU (Natural Language Understanding) module comprises three components: (i) the Logical Converter, (ii) the Logical Evaluator, and (iii) the Command Executor.
The human agent inputs commands or queries to the NLU module which undergo transformation via the Logical Converter and interpretation by the Logical Evaluator. 
Following sentence processing, either a command is sent to AbeSim for execution in the simulated environment, or the user is provided with information related to the environment state.

The Logical Converter performs the transformation of NL sentences into a form (see Section~\ref{section:logical-representation}) that can be interpreted by both the Logical Evaluator and the Command Executor. The conversion is achieved through a  GPT-3 Curie model fine-tuned by training on pairs of NL sentences and logical representations. The logical representations follow the syntax of Mace4. In case of a command, the Logical Evaluator determines whether it can be executed, and in case of a query, it returns a result (usually an integer or a boolean).

When the human agent sends a valid NL command to the system, after being processed by the Logical Converter and Evaluator, it reaches the Command Executor, which calls the AbeSim API based on the command type and the parameter list. Depending on the response from the API, the Executor updates the Mace4 expressions regarding the world state or returns an error.

\section{Logical Representation}
\label{section:logical-representation}

Every NL translation is structured as a JSON object. The following properties are part of the representation: 
(i) \texttt{type} - can take one of the 3 possible values: \texttt{"command"}, \texttt{"query"}, or \texttt{"invalid"} (the last value being used for a string that cannot be classified as either a command or a query);
(ii)  \texttt{expressions} - attribute included for both queries and commands, it contains a list of expressions to be evaluated;
(iii) \texttt{commands} - present only in case of commands.

This work aims to create a system that supports sentences containing a single command/query. For example, the clause ``Fetch \textit{7} carrots''  is valid, while the sentence ``Fetch \textit{7} carrots and cut \textit{them}'' cannot be processed because it contains two separate commands. Input of this nature poses a greater challenge as the model must discern that both commands in the sentence refer to the same objects (i.e. the 7 carrots). Handling such phrases may be part of a future improvement of the system. However,  for easier future extension, the ``expressions'' and ``commands'' properties are defined as lists, where each element in the list corresponds to a command/query in the input. In the current version, each list always contains only one item.

Listings~\ref{lst:command-ex1} and~\ref{lst:command-ex2} present the formalisation of two sample commands. Consider the first sentence: ``Cut 5 onions using a knife'' (Listing~\ref{lst:command-ex1}). Because the command contains 2 quantifiers, the ``expressions'' attribute contains in turn 2 elements:

\begin{enumerate}
      \item \texttt{"|exists x1 (onion(x1)).| >= 5"} - the first condition that must be fulfilled: there must be at least 5 onions in the kitchen. In addition to the restriction imposed on the objects, the quantifier indicates that the user wants the command applied to 5 objects;
      \item \texttt{"|exists x2 (cookingKnife(x2)).| >= 1"} - includes the second restriction: there must be at least one knife in the kitchen for the command to be executed.
\end{enumerate}

Note that both expressions contain the generalised quantifier \textit{at least}. 
Although this operator can be expressed in FOL (e.g. \textit{at least 2 O's} can be formalised with 
$\exists x\ \exists y\ O(x) \wedge O(y) \wedge x \neq y $), this representation becomes cumbersome for larger values. 
Hence, we opt to use the syntax above and parse the inner expressions (i.e. the formulas between vertical bars). 
Every inner expression is given to Mace4 to determine the cardinality of a variable (more details in  
Section~\ref{section:reasoning}).

The ``commands'' property is always written in classical FOL (i.e. cardinality is not required) and used to formalise and validate an action. For example, if the robot were asked to perform a prohibited or invalid action, such as cutting a bowl (``Cut a bowl''), through this logical representation, the execution of the command would be blocked.

\begin{lstlisting}[caption={Formalising the command ``Cut 5 onions using a knife''}, captionpos=b, label={lst:command-ex1}]
{ "type": "command", "expressions": [[ "|exists x1 (onion(x1)).| >= 5", "|exists x2 (cookingKnife(x2)).| >= 1" ]], "commands": [ "robot(x0) & onion(x1) & cookingKnife(x2) -> cut(x0, x1, x2)." ]}
\end{lstlisting}

Listing~\ref{lst:command-ex2} presents the representation of a command where an object is referenced by specifying the name assigned to it by the robot: ``Mix bowl LargeBowl1''. The ``expressions'' attribute contains an empty list because there are no cardinality restrictions in the sentence as a specific object is referenced, ``LargeBowl1'', of type ``bowl''. Additionally, even though not directly stated in the NL command, the predicate ``whisk(x2)'' is implicitly added to the logical representation of the command, since a whisk is also needed for performing the action.
The three types of commands are illustrated in Table~\ref{tab:commands-all}.

\begin{lstlisting}[caption={Formalising the command ``Mix bowl LargeBowl1''}, captionpos=b, label={lst:command-ex2}]
{ "type": "command", "expressions": [[]], "commands": [ "robot(x0) & bowl(LargeBowl1) & whisk(x2) -> mix(x0, LargeBowl1, x2)." ]}
\end{lstlisting}

\begin{table}
    \centering
    \caption{Three types of commands supported: quantifiers, referred objects and combination of these}
    \label{tab:commands-all}
\begin{tabular}{ll}
\hline
Command type & Example \\ \hline
Quantifiers & ``Fetch \textit{all} green peppers''  \\
& ``Cut \textit{several} bananas with a knife''  \\
& ``Cover \textit{3} trays with paper''  \\ 
Referred objects & ``Move contents of \textit{MediumBowl1} to \textit{MediumBowl2}''  \\ 
Quantifiers and referred objects & ``Next cut \textit{1} mango using cooking knife \textit{Knife1}'' \\ \hline
\end{tabular}
    
\end{table}

Listing~\ref{lst:query-ex1} contains the logical representation of a query (``Most vegetables are red onions'') using first-order logic with cardinality and following the syntax of Mace4. Additional examples of translations of NL to expressions in FOL with cardinality are included in Table~\ref{table:fol}.

\begin{lstlisting}[caption={Formalising the query ``Most vegetables are red onions''}, captionpos=b, label={lst:query-ex1}]
{"type": "query", "expressions": [ "|exists x0 (vegetable(x0) & redOnion(x0)).| > |exists x0 (vegdetable(x0) & -redOnion(x0)).|" ]}
\end{lstlisting}

Finally, Listing~\ref{lst:invalid-ex1} illustrates the general translation of an invalid or out-of-scope sentence, for example, ``I like swimming''.

\begin{lstlisting}[caption={Formalising the invalid sentence ``I like swimming''}, captionpos=b, label={lst:invalid-ex1}]
{ "type": "invalid" }
\end{lstlisting}

\begin{table*}
\caption{Examples of NL queries formalised in FOL with cardinality}
\label{table:fol}
\begin{tabular}{p{5cm}l}
\hline
Query               & FOL with cardinality   \\ \hline
All objects are boxes & \texttt{$\forall$x object(x) $\rightarrow$ box(x)} \\ 
No object is a box & \texttt{$\neg \exists$x object(x) $\wedge$ box(x)} \\ 
There is a box & \texttt{$\exists$x object(x) $\wedge$ box(x)} \\ 
There are at least two boxes              & \texttt{|$\exists$x box(x)| $\geq$ 2}           \\ 
There are exactly two boxes  &  \texttt{|$\exists$x box(x)| == 2}  \\ 
There are more boxes than tools & \texttt{|$\exists$x box(x)| > |$\exists$y tool(y)|} \\ 
Most objects are boxes                   & \makecell[l]{\texttt{|$\exists$x box(x) \& object(x)| > |$\exists$y} \\ \texttt{$\neg$box(y) \& object(y)|}}  \\ 
There are twice as many boxes as other objects & \makecell[l]{\texttt{|$\exists$x box(x)| == 2 $\times$ |$\exists$y $\neg$box(y) \&} \\ \texttt{object(y)|}} \\ 
There are many boxes & \texttt{|$\exists$x box(x)| $\geq threshold$}  \\ 
How many boxes are there? & \texttt{|$\exists$x box(x)|} \\  \hline
\end{tabular}
\end{table*}

\section{Supported Quantifiers}
\label{section:qunatifiers}

This paper focuses on English quantifiers with different levels of complexity, classified as ``natural language quantifiers of type ⟨1, 1⟩'' by Peters and Westerståhl~\cite{peters2006quantifiers}. 
There are two subcategories of NL quantifiers: (i) quantifiers with clear meaning, and (ii) quantifiers with subjective meaning. From the first category, this project handles the following quantifiers: \textit{all}, \textit{every}, \textit{none}, \textit{zero}, \textit{one}, \textit{two}, \ldots, $n$, \textit{at least one}, \textit{at least two}, \ldots, \textit{at least $n$}, \textit{at most one}, \textit{at most two}, \ldots, \textit{at most $n$}, \textit{exactly one}, \textit{exactly two}, \ldots, \textit{exactly $n$}, \textit{most}, \textit{majority of}, \textit{twice as many}, \textit{three times more}, \ldots, \textit{$n$ times more}, \textit{more than}, \textit{less than}, \textit{between $k_1$ and $k_2$}, \textit{a dozen}, \textit{half a dozen}, \textit{half of}, \textit{less than half of}, and \textit{more than half of}.

Table~\ref{table:amb} contains the ambiguous quantifiers handled in this project. Even if a hierarchy can be defined (e.g. MANY > SEVERAL > SOME > A FEW, as in~\cite{FEIMAN201629}), there are no specific values to be associated with these quantifiers. To address this issue, the GPT-3 model is trained to use the values in Table~\ref{table:amb} when one of the ambiguous quantifiers appears in the input given by the human agent. For example, if the command entered in the system is ``Cut several tomatoes``, the logical representation will contain the quantifier \textit{5} and the robot will cut 5 tomatoes. 
Another option to handle ambiguous quantifiers would be to ask for confirmation (``Are 5 tomatoes enough for you?'') or to inform the user about the current interpretation of the quantifier (``I will cut five tomatoes'').

\begin{table}
\centering
\caption{\label{table:amb}Examples of numerical values associated with ambiguous quantifiers}
\begin{tabular}{l|cccccc}
\hline
Quantifier & A couple & Few & A few & Some & Several & Many\\ \hline
Value & 2 & 2 & 3& 4 & 5 & 10 \\ \hline
\end{tabular}

\end{table}

\section{Reasoning and Evaluating FOL Expressions}
\label{section:reasoning}

\subsection{State Representation and Background Knowledge}

The NLU module obtains the world representation (i.e. the existing object types and assigned names) from AbeSim  and automatically populates the file ``\texttt{sensors.in}'' (see Listing~\ref{lst:sensorexemp}) with the domain size (i.e. the total number of objects in the simulation), a distinction list, and atomic statements such as ``\texttt{tomato(Tomato1)}'' which indicates that there is an object of type ``\texttt{tomato}''. Every object is assigned a constant (i.e. a distinct function with values between 0 and \texttt{domain\_size - 1} through the use of ``\texttt{list(distinct)}''). For example, in Listing~\ref{lst:sensorexemp}, constant ``\texttt{Tomato1}'' is assigned function \texttt{1}.  Without this detail, Mace4 would also generate models (interpretations) in which the same function is associated with constants having different names, assuming that there are cases when they refer to the same physical object.

\begin{lstlisting}[caption={Sample content for file ``\texttt{sensors.in}''}, captionpos=b, label={lst:sensorexemp}]
assign(domain_size, 5).

list(distinct).
    [Robot1, Tomato1, Tomato2, Whisk1, CookingKnife1].
end_of_list.

formulas(sensors).
    robot(Robot1). tomato(Tomato1). tomato(Tomato2). 
    whisk(Whisk1). cookingKnife(CookingKnife1).
end_of_list.
\end{lstlisting}

While formulas in ``\texttt{sensors.in}'' are dynamically created and depend on the kitchen state, the content of the file ``\texttt{background\_knowledge.in}'' is static as it contains the ontology of the system. 
It consists of 3 types of expressions: (i) \textit{classification}, (ii) \textit{distinction}, and (iii) \textit{commands}. The first 2 types of formulas establish the relations between different types (predicates) of objects using FOL implication (e.g. in Listing~\ref{lst:backgexemp}, both ``\texttt{cookingKnife}'' and ``\texttt{whisk}'' objects are classified as a ``\texttt{kitchenTool}'', but a ``\texttt{cookingKnife}'' cannot be a ``\texttt{whisk}'' too). In contrast, the \textit{commands} expressions declare the possible commands that can be executed by a robot and define some restrictions (e.g. in Listing~\ref{lst:backgexemp}, it is imposed that a robot cannot fetch another robot or that only robots can fetch other objects).

\begin{lstlisting}[caption={Sample content for file ``\texttt{background\_knowledge.in}''}, captionpos=b, label={lst:backgexemp}]
formulas(background_knowledge_classification).
    tomato(x) -> ingredient(x). 
    cookingKnife(x) -> kitchenTool(x). whisk(x) -> kitchenTool(x).
end_of_list.

formulas(background_knowledge_distinction).
    ingredient(x) | kitchenTool(x) -> -robot(x).
    robot(x) | kitchenTool(x) -> -ingredient(x).
    robot(x) | ingredient(x) -> -kitchenTool(x).
    cookingKnife(x) -> -whisk(x).
end_of_list.

formulas(background_knowledge_commands).
    robot(x) & (ingredient(y) | kitchenTool(y)) -> fetch(x, y).
    -robot(x) -> -fetch(x, y).
    -ingredient(y) & -kitchenTool(y) -> -fetch(x, y).
end_of_list.
\end{lstlisting}

\subsection{Generating Interpretation Models}
The finite models generator Mace4 is used to compute all interpretations of an expression.
Expressions that require evaluation are added to the file ``\texttt{expression.in}''. 
Consider Listing~\ref{lst:exprcqexemp} which contains an expression used for verifying whether there is at least one ingredient in the kitchen. 
By running Mace4 against all 3 files (Listings~\ref{lst:sensorexemp}, \ref{lst:backgexemp}, and \ref{lst:exprcqexemp}), we obtain 2 models that satisfy the formulas (i.e. the quantified variable \texttt{x} from ``\texttt{expression.in}'' can take only 2 values, ``\texttt{Tomato1}'' or ``\texttt{Tomato2}'', the only constants for which the predicate ``\texttt{ingredient}'' is true). 
The returned models have dual importance: they can be used
(1) to determine the cardinality of a predicate; and 
(2 to obtain a list with objects that satisfy the given conditions. 

\begin{lstlisting}[caption={Sample content for file ``\texttt{expression.in}''}, captionpos=b, label={lst:exprcqexemp}]
formulas(expressions).
    exists x (ingredient(x)).
end_of_list.
\end{lstlisting}

\begin{figure}
\centering
\includegraphics[width=0.5\textwidth]{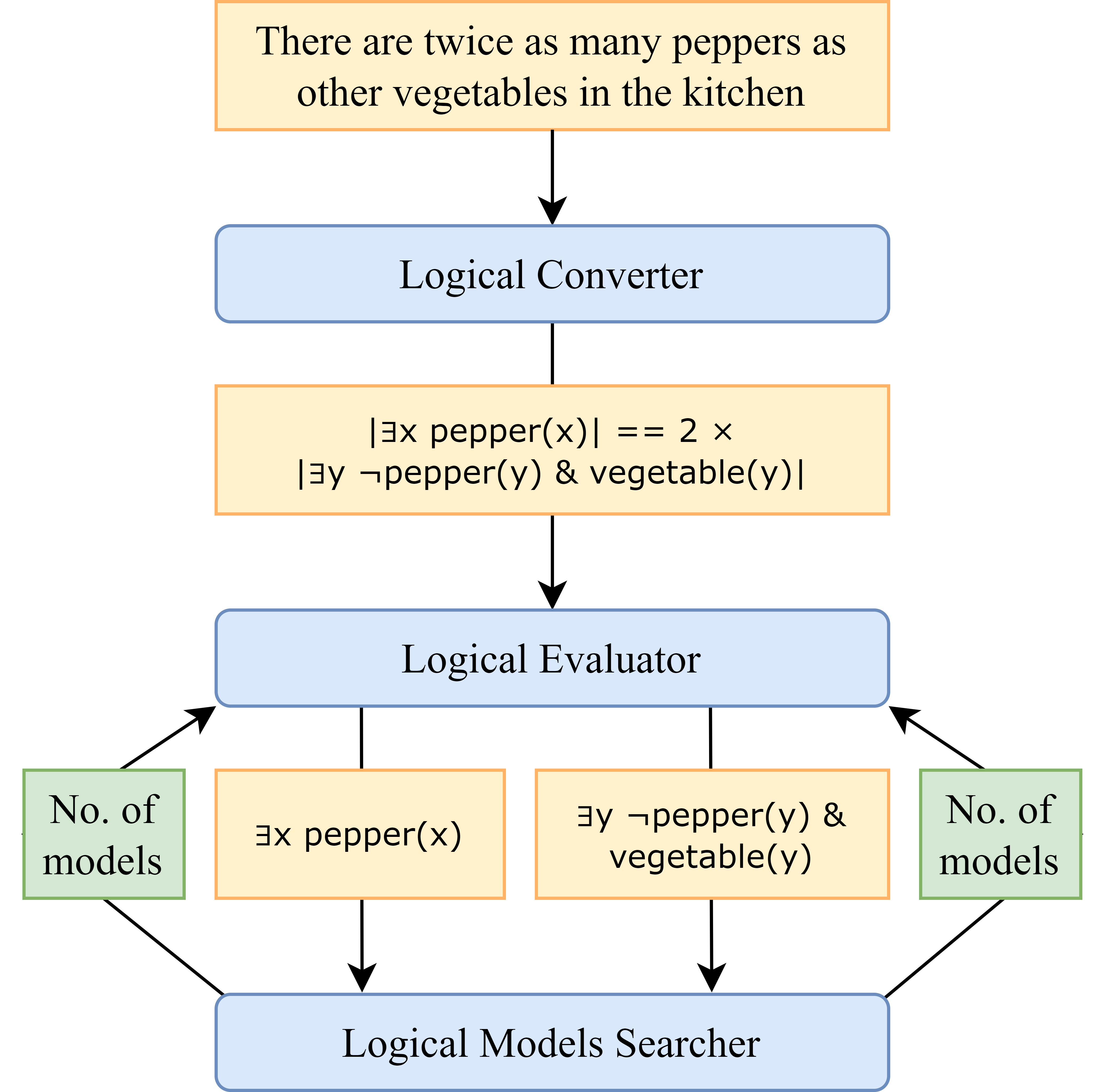}
\caption{Execution flow for a query}
\label{fig:flow}
\end{figure}

One example of a complete interpretation flow of a query with cardinality appears in Figure~\ref{fig:flow}. 
For the sentence ``There are twice as many peppers than other vegetables in the kitchen'', the Logical Converter obtains the following translation from GPT-3: ``\texttt{|exists x (pepper(x)).| == 2 * |exists y (-pepper(y) \& vegetable(y)).|}''. The two expressions between vertical bars are extracted by the Logical Evaluator, which evaluates them individually. For each execution, the Evaluator reads the number of models returned by Mace4, thereby determining the cardinality for each type of object in the initial expression. Assuming that 10 models are returned for the formula ``\texttt{|exists x (pepper(x)).|}'' and only 5 for ``\texttt{|exists y (-pepper(y) \& vegetable(y)).|}'', the evaluator concludes that the statement is true.

\section{AbeSim Simulator}
\label{section:abe}

\begin{figure}
\centering
\includegraphics[width=0.37\textwidth]{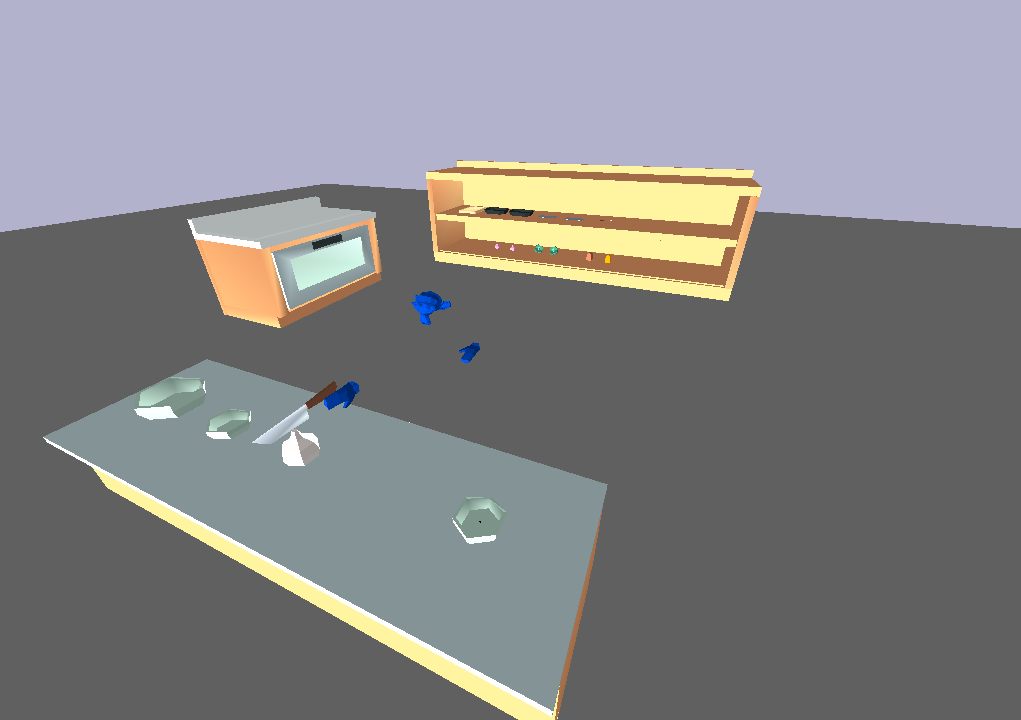}
\caption{Execution of the ``to-cut'' HTTP call in the AbeSim simulator \cite{mpomarlan/abe_sim}}
\label{fig:abesim}
\end{figure}

AbeSim\footnote{\url{https://github.com/mpomarlan/abe_sim}} is a simulation environment built on top of the PyBullet\footnote{\url{https://pybullet.org/wordpress/}} robotics simulator (see Figure~\ref{fig:abesim}). AbeSim is developed to function as a testing component for an NLU benchmark to be run by the University of Brussels~\cite{de_haes2023benchmark}. As part of this benchmark, an NLU software would have to analyse a cooking recipe and convert it into a meaning representation. The correctness of the meaning representation is tested by converting it into a series of primitive action calls that an ``agent'' in the AbeSim world can execute. 

Therefore, the focus of AbeSim is not on robotics details; the agent has two hands that float without arms (i.e. depicted with blue in Figure~\ref{fig:abesim}) and have a wide reach, which simplifies kinematics and planning. 
Instead, AbeSim focuses on providing an environment with a wide variety of objects that the agent can interact with, and where the objects provide various affordances beyond what rigid body simulation allows. As such, some objects can generate others, e.g. it is possible to pour various substances out of appropriate bags or bottles; some objects represent food items that can be peeled, cut, mashed, etc. when tools of appropriate types perform certain motions; objects have a temperature property that updates based on their surroundings in a very rough approximation of heat transfer. Note, however, that all such interactions, including cutting, mashing etc., are modelled via scripts that trigger on appropriate motions, not physics simulation.

The AbeSim agent (``Abe'') can be communicated with via a HTTP POST interface, which defines a set of commands such as going towards a given object, placing a given object at a particular location, and performing one basic operation for a cooking recipe, such as cutting, baking, etc. The HTTP POST interface also allows querying the world state. This state consists mostly in numeric data, in particular the values of physics simulation-related variables such as positions, velocities and orientations for each object, as well as ``custom state variables'' that pertain to the objects' other affordances, e.g. what temperature an object is at, how close it is to being peeled and so on. However, for each object, a symbolic location property ``at'' is provided. This assists in reconstructing a qualitative description of some of the spatial relations between the objects in the scene. For example, a cup placed on the table is ``at'' the table, and coffee particles in the cup are ``at'' the cup.

\section{Fine-tuning GPT-3 for translating NL to FOL}
\label{section:fine-tuning}
Here, we present the dataset utilised for fine-tuning GPT-3, along with the corresponding experimental results obtained from the fine-tuned model.
\subsection{Dataset for fine-tuning Curie GPT-3}

The dataset used to fine-tune the Curie GPT-3 model contains pairs of NL sentences (\textit{prompts}) and the corresponding translations (\textit{completions}) into FOL. All translations follow the structure described in Section~\ref{section:logical-representation} (see Listings~\ref{lst:command-ex1},~\ref{lst:command-ex2},~\ref{lst:query-ex1}, and~\ref{lst:invalid-ex1}).

Since there are three types of input sentences (i.e. commands, queries, and invalid statements), the dataset includes examples for each category (see Table~\ref{tab:datasetoveralldistribution}). 
The training set contains 396 examples (60\%), while the validation and the test data sets comprise 132 examples each (20\%).



The distribution of examples based on the corresponding command to be executed appears in Table~\ref{tab:datasetcommanddistribution}.
For instance, in the training dataset, there are 59 translations for the ``fetch'' command and only 16 for the ``transfer'' command. 
This distribution reflects the varying ways in which a command can be used: ``fetch'' has a larger scope, various quantifiers or object names can be attached to it (e.g. ``Fetch \textit{several} / \textit{many} / \textit{a couple} / \textit{at least 5 onions} / \textit{LargeBowl1}''). 
In contrast, ``transfer'' is generally used to move the content from one specific container to another (e.g. ``Transfer contents of \textit{MediumBowl1} to \textit{another} bowl''). .  

\begin{table}
\parbox{.25\linewidth}{
\parbox{.25\linewidth}{
\centering
\caption{Dataset distribution \label{tab:datasetoveralldistribution}}    
\begin{tabular}{lccc}
\hline
 Pair type             & Train & Dev & Test \\   \hline
 Command          &  240  &  80        &  80     \\ 
 Query            &  96   &  32        &  32     \\ 
 Invalid          &  60   &  20      &  20     \\  \hline
 Total            &  396 (60\%)  &  132 (20\%)    &  132 (20\%)   \\  \hline
\end{tabular}
}
\\ \vspace{8mm}
\\
\parbox{.25\linewidth}{
\centering
\caption{Command distribution in the dataset \label{tab:datasetcommanddistribution}}    
\centering
\begin{tabular}{lccc}
\hline
 Command            & Train & Dev & Test   \\   \hline
 fetch     &  59       &  19      &  19       \\ 
 cut       &  38       &  12      &  12       \\ 
 bake      &  36       &  12      &  12       \\ 
 line      &  36       &  12      &  12       \\ 
 mix       &  26       &  9       &  9        \\ 
 transfer  &  16       &  6       &  6        \\  
 sprinkle  &  16       &  6       &  6        \\ 
 shape     &  13       &  4       &  4        \\   \hline
 Total     &  240      &  80      &  80       \\ \hline
\end{tabular}
}}
\hfill
\parbox{.44\linewidth}{
\centering
\caption{Query distribution by quantifier\label{tab:datasetquerydistribution}}
\begin{tabular}{lccc}
\hline
Query                & Train  & Dev   & Test \\ \hline
most/majority of              & 6          & 3          &   3      \\ 
more than                     & 6          & 2          &   3      \\ 
less than                     & 5          & 1          &   2      \\
at most                       & 4          & 1          &   1      \\
at least                      & 3          & 1          &   1      \\
exactly/only                  & 7          & 2          &   1      \\
$n$                           & 3          & 1          &   2      \\
$n$ times more                & 4          & 2          &   1      \\
between $k_1$ and $k_2$       & 4          & 1          &   1      \\
many/a lot                    & 4          & 2          &   1      \\
several                       & 3          & 1          &   1      \\
a few/few                     & 5          & 2          &   2      \\
a couple                      & 3          & 1          &   1      \\
some                          & 4          & 1          &   1      \\
how many/count                & 7          & 2          &   2      \\
half                          & 3          & 1          &   2      \\
no/none                       & 4          & 2          &   1      \\
all/every                     & 8          & 3          &   3      \\
dozen/half a dozen            & 4          & 1          &   1      \\
\textit{combinations}         & 9          & 2          &   2       \\  \hline
Total                         & 96         & 32         & 32     \\ \hline
\end{tabular}
}
\end{table}

Table~\ref{tab:datasetquerydistribution} lists the number of queries included in the dataset distributed based on the quantifier type. 
There are 96 queries for training, 32 for validation and 32 for testing. Some examples for this type of sentences are ``There are \textit{several} doughnuts'', ``There are \textit{twice as many} onions than carrots'' or ``\textit{How many} green chili peppers are there in the kitchen?''. Note that multiple quantifiers can occur in the same sentence, e.g. ``There are \textit{less than} \textit{a dozen} eggs''

Besides commands and queries, the dataset contains 60 examples of invalid or out-of-scope input for the investigation domain. It is important to train the model to identify commands that the robot is not able to process. 
Some examples are: ``Transfer nothing'', ``Cut something'', ``Recommend me a movie''. 
The performance of the translation model is assessed using 20 examples of invalid statements for validation and other 20 for testing. 

For fine-tuning the Curie model~\cite{finetuning},~\cite{finetuningapi}, we used the following parameters:
(i) \textit{n\_epochs} = 4, the default value; 
(ii) \textit{batch\_size} = 4, different from the default value which is 0.2\% from the total number of training examples, that would be in our case 0.2\% $\times$ 396 = 0.792;
(iii) \textit{learning\_rate\_multiplier}, default value (0.05, 0.1, or 0.2), automatically set based on \textit{batch\_size}.
The entire fine-tuning process lasted approximately 8 minutes, including 4 minutes for waiting in the OpenAI queue. The remaining 4 minutes were used for the actual training process (i.e. 1 minute for each epoch). The total fine-tuning cost was \$0,28.

\subsection{Assessing the quality of translation}
\label{section:results}

OpenAI provides the following metrics for the analysis of the fine-tuning results~\cite{finetuning}: 
(i) loss;
(ii) sequence accuracy; and
(iii) token accuracy.
While the sequence accuracy represents the percentage of completions for which the model generated an output identical to the expected one, the token accuracy is more flexible as it represents the percentage of tokens that were generated as expected, where a token is a ``piece of word'', ``about 4 chars''~\cite{tokenhelp} in the completion. 

When it comes to the task of converting NL into logical representation, it is not imperative for a generated completion to be identical to the expected one in order for the translation to be correct. For instance, the obtained translation might include an extra leading space for one of the tokens. The presence of the leading space does not affect the correctness of the expression; therefore, even if the two completions are not identical, they are equivalent. In this context, the sequence accuracy metric can be considered too rigid for our semantic reasoning task.

\begin{figure}
\centering
\begin{subfigure}{0.5\textwidth}
  \centering
  \includegraphics[width=\textwidth]{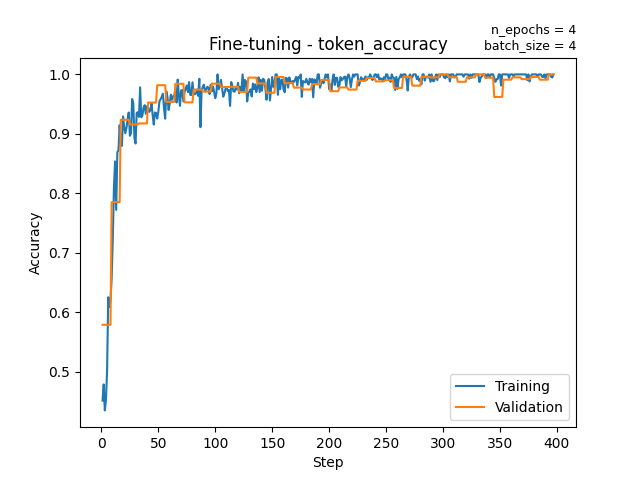}
  \caption{Token accuracy}
  \label{fig:accuracy}
\end{subfigure}%
\begin{subfigure}{0.5\textwidth}
  \centering
  \includegraphics[width=\textwidth]{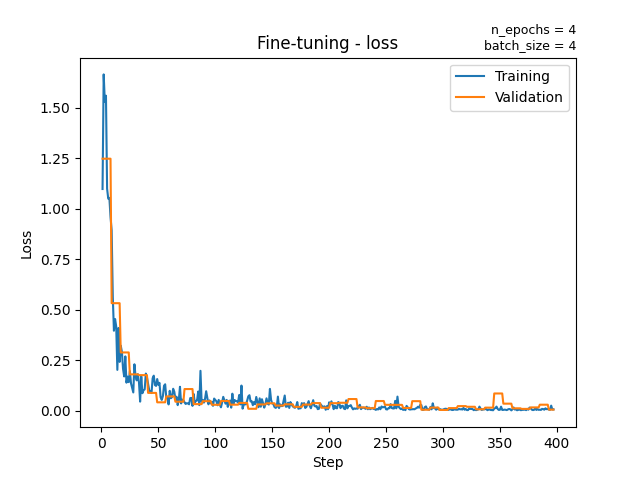}
  \caption{Loss}
  \label{fig:loss}
\end{subfigure}
\caption{Training and validation token accuracy and loss}
\label{fig:acurracy-loss}
\end{figure}

The token accuracy represents a metric that helps us assess the performance of the model. 
Figures~\ref{fig:accuracy} and \ref{fig:loss} depict the comparative evolution of token accuracy for both training and validation datasets, as well as the corresponding training and validation loss. The accuracy for the training set in the final step is 99.48\%, while for the validation set, it is 99.11\%. However, the large accuracy values do not reflect the true performance of the model as the tokens that were predicted incorrectly for the completions might be essential (i.e. the absence of a parenthesis).

Because sequence accuracy is a too strict as a metric for assessing the model accuracy (different logical expressions may be equivalent), an additional step was introduced in validation: whenever the model generated a completion different from the one in the validation set, the completion was manually verified. Based on this approach, the following results can be outlined:  
\begin{itemize}
\item 97 generated completions (73.48\%) were identical to the expected ones;
\item 9 completions (6.81\%) were correct but with a different form than the expected one;
\item 26 completions (19.69\%) were different in meaning than the one expected.
\end{itemize}

Thus, the practical accuracy for the validation set (97 identical plus 9 logical equivalent) is $(97 + 9) / 132 = 80.30\%$.

To control the level of randomness in the obtained output, the ``temperature'' parameter~\cite{modelsdocumentation} was considered. 
It takes values ranging from 0 to 2, with a default setting of 1.
For a value of 0, the output of the model is deterministic (i.e. two calls to the model with the same prompt will always generate the same response). 
We experimented with three values: 0, 0.5, and 1. 
During the analysis of the results, it was observed that the parameter yielded the highest number of accurately generated completions when set to 0. 

Recall that  6.81\% of the translations deviated from the expected form, but their representation was still correct. 
To have an understanding of these cases, where the model generates a correct completion but with different tokens, 
we can analyse the translation for the given prompt: ``Blend the contents of the bowl \textit{Bowl1} using a whisk''. 
The expected translation and the generated one appear in Listing~\ref{lst:blenddiff}. 
The only difference between the two completions is the variable name used to check for the presence of a whisk: the completion from the validation set (i.e. the expected one) uses the variable ``x2'', whereas the generated completion uses the variable ``x1''. 
The variable name does not affect the outcome, as the logical expressions are equivalent.

\begin{lstlisting}[caption={A different but correct translation for the prompt: ``Blend the contents of the bowl Bowl1 using a whisk''}, captionpos=b, label={lst:blenddiff}]
# Expected
{'type':'command','expressions':[['|exists x2 (whisk(x2)).| >= 1']], 'commands':['robot(x0) & bowl(Bowl1) & whisk(x2) -> mix(x0, Bowl1, x2).']} 
# Generated
{'type':'command','expressions':[['|exists x1 (whisk(x1)).| >= 1']], 'commands':['robot(x0) & bowl(Bowl1) & whisk(x1) -> mix(x0, Bowl1, x1).']}
\end{lstlisting}

An incorrect translation is presented in Listing~\ref{lst:expressiondiff} for the prompt: ``All peppers are not red peppers''. 
This is a statement about the kitchen state, which should be confirmed or denied by the system. 
There is only a tiny but significant difference between the correct output and the generated one: the negation operator is not present before the ``\texttt{redPepper}'' predicate. 
Hence, instead of the correct query, the system answers ``yes'' if all peppers are red peppers. 
\begin{center}
\begin{lstlisting}[caption=Incorrect translation for the prompt ``All peppers are not red chili peppers'', captionpos=b, label={lst:expressiondiff}] 
# Expected
{'type':'query','expressions':['all x0 (pepper(x0) -> -redPepper(x0)).']}
# Generated
{'type':'query','expressions':['all x0 (pepper(x0) -> redPepper(x0)).']} 
\end{lstlisting}
\end{center}

By analysing the errors obtained on the validation set, we extended the training set with more prompt-completion pairs considering the situations where the model had difficulties to create a correct completion. 
By applying fine-tuning with the consolidated training set of 396 pairs, the accuracy on the validation dataset was increased to the mentioned 80,30\%. 

The final evaluation of the model was performed using the test dataset, which consisted of prompts that were unknown to the model.
The methodology employed was identical to the additional validation stage: the model was invoked using the 132 prompts from the test set, and the completions generated by the model were compared to those in the test set. 
The translations that differed from the expected ones were individually analysed to determine if they would produce the expected result.
The obtained results on the test dataset were as follows:
\begin{itemize}
\item 93 translations (70,45\%) were identical to the expected ones;  
\item 7 translations (5,30\%) had small variations, but they were logically equivalent; 
\item 32 translations (24,24\%) were wrong, leading to a different interpretation. 
\end{itemize}
Hence, the practical accuracy of the model on the test dataset is $(93 + 7) / 132 = 75,75\%$.  

\section{Related Work}
\label{section:related-work}


Liu et al. have surveyed semantic reasoning techniques for robotic applications~\cite{LIU2023104294}. 
They have identified several ``computational frameworks'', such as \textit{semantic graphs}, \textit{Markov} or \textit{Bayesian networks}, \textit{description logics}, or \textit{first-order probabilistic models}. All computational frameworks are assessed on several characteristics, with first-order probabilistic models being described as effective in terms of \textit{modeling uncertainty} and \textit{knowledge representation adaptability}, but expensive in terms of \textit{expressiveness} (the complexity of the representation) and \textit{scalability}.

Darvish et al.~\cite{9257394} have proposed a ``hierarchical human-robot cooperation architecture'', divided into three layers: \textit{perception}, \textit{representation}, and \textit{action}. At the representation level, they have used FOL combined with \textit{AND/OR graphs} mentioning that the main advantage of this approach is the decoupling between the assertions about objects and the cooperation process.

Lemaignan et al.~\cite{LEMAIGNAN201745} also have used FOL statements as communication means between the components of system. 
Similarly to our implementation, their system contains a knowledge base of FOL statements that are queried by a language processing module, while an execution controller is triggered by certain assertions. 
Luckchuck et al.~\cite{luckcuck2022compositional} have used FOL to verify robot behaviour in a system composed of nodes which communicate via messages. Each node is specialised on a certain task, its specification being expressed in FOL.
Telx et al.~\cite{doi:10.1146/annurev-control-101119-071628} have outlined the approaches for handling NL in robotics, including learning mappings between NL and a formal language; they have presented several datasets used in language grounding.


To formalise quantifiers, several researchers~\cite{cui-etal-2022-generalized-quantifiers},~\cite{inproceedings},~\cite{mineshima-etal-2015-higher}, have focused on generalised quantifiers (GQs), which are derived from the existential ($\exists$) and universal ($\forall$) quantifiers of FOL, but extended over sets~\cite{cui-etal-2022-generalized-quantifiers}. 
Mineshima et al.~\cite{mineshima-etal-2015-higher} have proposed a higher-order inference system based on GQs. 
For example, the representation of \textit{Most students work} is ``$most$($\lambda x.student(x),\,\lambda x.work(x)$)''~\cite{mineshima-etal-2015-higher}, where ``$most$()'' is a higher-order predicate with first-order predicates as arguments. 
Pauw and Spranger~\cite{inproceedings} have studied the applicability of GQs to robot interaction scenarios by comparing them to an original approach called \textit{clustering determination}. 
Their work analyses the use of determiners, such as \textit{all} or \textit{the}, in noun phrases. They have applied a scoring mechanism to determine the objects which are part of a certain category, exemplifying the technique for the phrase ``the left block'': the occurrence of determiner ``the'' indicates that only one object should be selected (i.e. the one with the highest score) from all objects that belong to both class ``block`` and category ``left'' (implemented as a direction vector). 
Bott et al.~\cite{10.1093/jos/ffy015} have focused on determining quantifier scope by developing a ``cognitively realistic'' theory using empty-set effects. For instance, their representation of ``most'' is based on sets: \textit{Most dots are blue} becomes  ``$|dot \cap blue| > |dot - (dot \cap blue)|$''.


\section{Conclusion}
\label{section:conclusion}

We introduced a method for formalising quantifiers based on first-order logic, extended with variable cardinality. 
We applied the method on a human-robot interaction scenarios, i.e. kitchen domain. 
Two types of sentences are supported: (i) queries, and (ii) commands. We presented two main parts: the translation of natural language to logic and the process of reasoning against a knowledge base.

The ongoing work regards the contextual inference of objects referred to when using pronouns, based on previously processed sentences: e.g. when the human agent sequentially enters the command ``Fetch a red pepper'' and ``Cut it'', the robot should understand that in the second sentence, the object being referred to is the one brought after executing the first command. 
Other improvements can be: adjusting the representation of ambiguous quantifiers for a more flexible interpretation (i.e. using intervals instead of fixed values), introducing new commands and objects, or using a different base model from OpenAI for fine-tuning. As reference, the estimated effort for writing 50 prompt-completion pairs to train the AI model to learn a new command is about 1.5-2 hours. Davinci, the most capable GPT-3 model, has the potential to provide better accuracy compared to the current Currie model, which has an accuracy of 75.75\%. However, this improvement may come at the cost of increased generation time for translations.

\bibliography{main}

\appendix

\end{document}